%% file: main.tex
\documentclass[10pt,twocolumn,letterpaper]{article}

\usepackage[pagenumbers]{cvpr} 

\input{preamble}

\definecolor{cvprblue}{rgb}{0.21,0.49,0.74}
\usepackage[pagebackref,breaklinks,colorlinks,citecolor=cvprblue]{hyperref}

\def\confName{CVPR}
\def\confYear{2024}

\title{FFF: Fixing Flawed Foundations in contrastive pre-training results in very strong Vision-Language models}

\author{Adrian Bulat$^{1,2}$ \quad Yassine Ouali$^{1}$ \quad Georgios Tzimiropoulos$^{1,3}$ \vspace{0.3em} \\
{\normalsize $^1$Samsung AI Center Cambridge, UK} \quad
{\normalsize $^2$ Technical University of Iași,  Romania} \quad \\
{\normalsize $^3$ Queen Mary University of London, UK}
}

\begin{document}
\maketitle
\input{sec/0_abstract}    
\input{sec/1_intro}
\input{sec/2_related_work}
\input{sec/3_method}

\input{sec/4_results}
\input{sec/5_ablations_and_conclusion}
\clearpage
{
    \small
    \bibliographystyle{ieeenat_fullname}
    \bibliography{main}
}


\input{sec/X_suppl}

\end{document}

%% file: preamble.tex
\usepackage{environ}         
\usepackage{etoolbox}        
\usepackage{graphicx}        

\newlength{\myl}
\let\origequation=\equation
\let\origendequation=\endequation

\RenewEnviron{equation}{
  \settowidth{\myl}{$\BODY$}                       
  \origequation
  \ifdimcomp{\the\linewidth}{>}{\the\myl}
  {\ensuremath{\BODY}}                             
  {\resizebox{\linewidth}{!}{\ensuremath{\BODY}}}  
  \origendequation
}

\RenewEnviron{equation*}{
  \settowidth{\myl}{$\BODY$}                       
  \ifdimcomp{\the\linewidth}{>}{\the\myl}
  {\ensuremath{\BODY}}                             
  {\resizebox{\linewidth}{!}{\ensuremath{\BODY}}}  
}

\usepackage{amssymb}
\usepackage{pifont}

\usepackage[dvipsnames]{xcolor}

\newcommand{\xmark}{\ding{55}}%

%% file: sec/0_abstract.tex
\begin{abstract}

Despite noise and caption quality having been acknowledged as important factors impacting vision-language contrastive pre-training, in this paper, we show that the full potential of improving the training process by addressing such issues is yet to be realized. Specifically, we firstly study and analyze two issues affecting training: incorrect assignment of negative pairs, and low caption quality and diversity. Then, we devise effective solutions for addressing both problems, which essentially require training with multiple true positive pairs. Finally, we propose training with sigmoid loss to address such a requirement. We show very large gains over the current state-of-the-art for both image recognition ($\sim +6\%$ on average over 11 datasets) and image retrieval ($\sim +19\%$ on Flickr30k and $\sim +15\%$ on MSCOCO).

\end{abstract}

%% file: sec/1_intro.tex
\section{Introduction}
\label{sec:intro}

Large-scale contrastive image-text pre-training has emerged as the prevalent method for vision-language representation learning~\cite{radford2021learning, jia2021scaling,li2021supervision,li2022blip,yao2021filip,you2022learning,furst2022cloob}. The majority of datasets employed for pre-training are web-collected~\cite{sharma2018conceptual, changpinyo2021cc12m, desai2021redcaps, srinivasan2021wit, nguyen2022quality, thomee2016yfcc100m, schuhmann2021laion, schuhmann2022laion}. They offer a varied data distribution and are sufficiently large to effectively train high-performing vision-language models.
However, since the raw captions for each image are typically extracted from associated tags
or descriptions, they often exhibit low quality, being noisy and suboptimal for training purposes~\cite{jia2021scaling,li2022blip}. Although some attempts to fix such issues have been already described, to some extent, in literature (e.g. ALIP~\cite{yang2023alip}, BLIP~\cite{li2022blip}), in this work, we show that the full potential of improving the quality of the training process is far from being fully realized. Specifically, by studying and addressing specific issues related to noise and low data quality, in this work, we show that our improved vision-language training pipeline can achieve massive gains over the current state-of-the-art methods for both image recognition ($\sim +6\%$ on average over 11 datasets) and image retrieval ($\sim +19\%$ on Flickr30k~\cite{young2014image} and $\sim +15\%$ on MSCOCO~\cite{lin2014microsoft}).

The first issue we study is related to noise impacting contrastive learning: near-duplicate samples which are incorrectly treated as negative pairs. Even within a batch, it is not uncommon to find images and/or captions that are semantically similar or even identical. Since standard contrastive learning assumes one positive pair, this significantly hinders the training process and the quality of the trained models. 

\begin{figure}
    \includegraphics[width=0.45\textwidth]{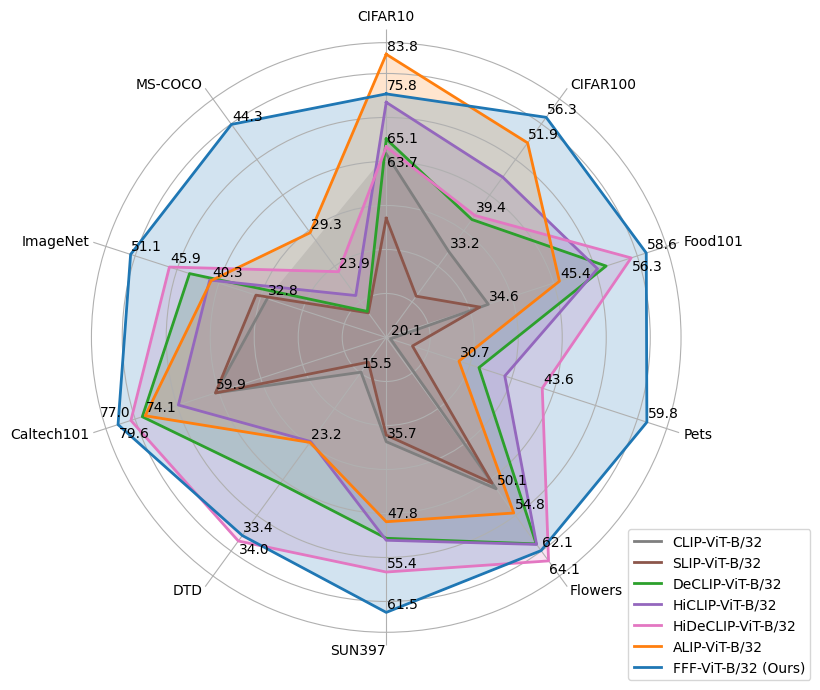}
    \caption{Our approach, \textbf{FFF}, achieves state-of-the-art accuracy across multiple datasets, largely outperforming prior methods.}
    \vspace{-0.3cm}
\end{figure}

The second issue we study is related to low caption quality and diversity. Captions can be short and lacking detail, noisy,  or even entirely irrelevant to the image. Moreover, since the mapping process between image and text is one-to-many, more than one caption is needed to provide an approximate description of the image.

To fix issue one, we propose an algorithm that mines new positive pairs based on image-text, image-image, and text-text similarities, aiming to decrease the number of false
negatives in the training data arising due to semantically similar images and/or captions. 

We fix issue two by firstly generating pseudo-captions for each training image using a state-of-the-art image captioning technique~\cite{li2023blip} that will act as new true positives for a given image. Then, we propose \textit{batch text augmentation} for training with multiple pseudo-captions (\ie five captions per image selected via beam search) within the same batch to effectively increase caption diversity. 

Importantly, after applying the proposed solutions, we end up with a variable number of positive pairs per image \ie newly mined positive pairs and multiple pseudo-captions per image. This implies that we need to train our model with a loss function that accommodates multiple positives and is robust to potential errors in the mining process. Unfortunately, neither contrastive loss~\cite{radford2021learning} nor supervised contrastive loss~\cite{khosla2020supervised} can be directly applied for this case. To this end, we propose to use the sigmoid loss~\cite{zhai2023sigmoid} which allows the number of positives to vary dynamically per sample and per batch at no extra cost and is also robust to noise.

\begin{figure*}
    \centering
    \includegraphics[width=0.8\textwidth]{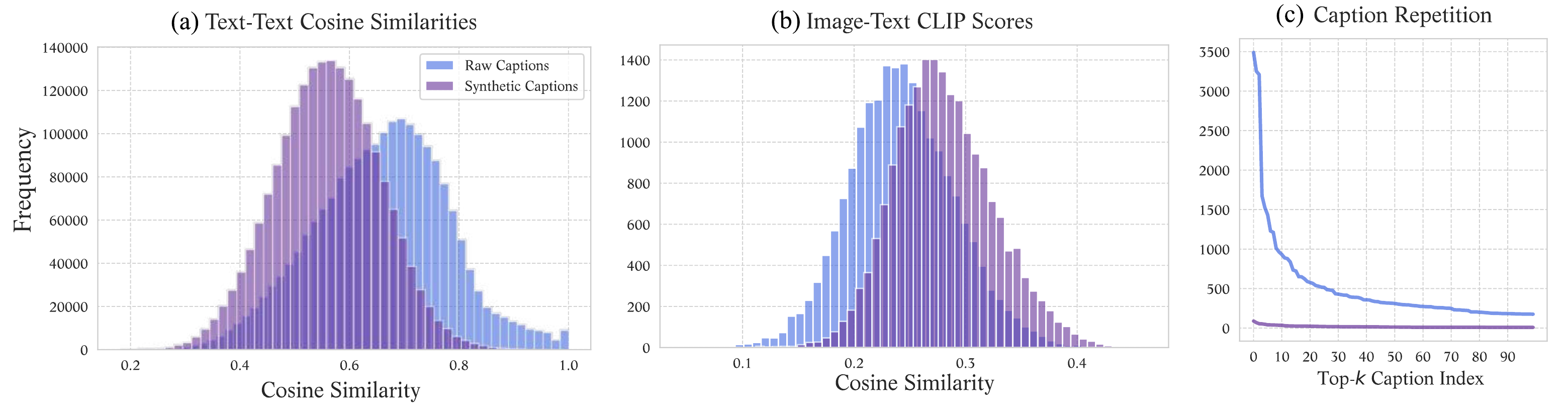}
    \caption{\textbf{Semantic and lexical diversity of raw and synthetic pseudo-captions of CC3M:}
    (a): Average cosine similarities of each caption and its 100 most similar captions using CLIP ViT-L/14 features.
    (b): Cosine similarities between features of each image and its ground-truth caption.
    (c): The frequencies of the top-100 most frequent raw and synthetic pseudo-captions (generated using BLIP2).
    We observe that the raw captions are semantically similar to each other (a), often not well aligned with their
    associated ground-truth images (b), and contain a high number of basic and redundant captions (c).
    By swapping them with  pseudo-captions, we observe an improved diversity (a,c) and better image-text alignment (b).}
    \label{fig:fig1}
\end{figure*}

Overall, we make the following \textbf{contributions}: 
\begin{itemize}
    \item We study and provide in-depth analyses of two important issues related to vision-language training process/data: false negative pairs due to semantic near-duplicates, and low caption quality and diversity (Sec.~\ref{sec:analysis}).
    \item
    We provide two simple algorithms for addressing the aforementioned issues: The first one uses text-image, image-image, and text-text similarities for eliminating incorrectly assigned negatives and mining new true positives. The second uses the proposed \textit{batch text augmentation} for training with multiple pseudo-captions per image within the same batch. Both solutions induce multiple new positives per each training image. To address this, we propose to use sigmoid loss for training the model. See Sec.~\ref{sec:method}.  
    \item 
    We show very large gains over the current state-of-the-art for both image recognition ($\sim +6\%$ on average over 11 datasets) and image retrieval ($\sim +19\%$ on Flickr30k and $\sim +15\%$ on MSCOCO) (Sec.~\ref{sec:results}). We further ablate the impact of many important components of our method in Sec.~\ref{sec:ablation_studies}.
\end{itemize}

\begin{figure*}[ht!]
    \centering
    \includegraphics[width=0.95\textwidth]{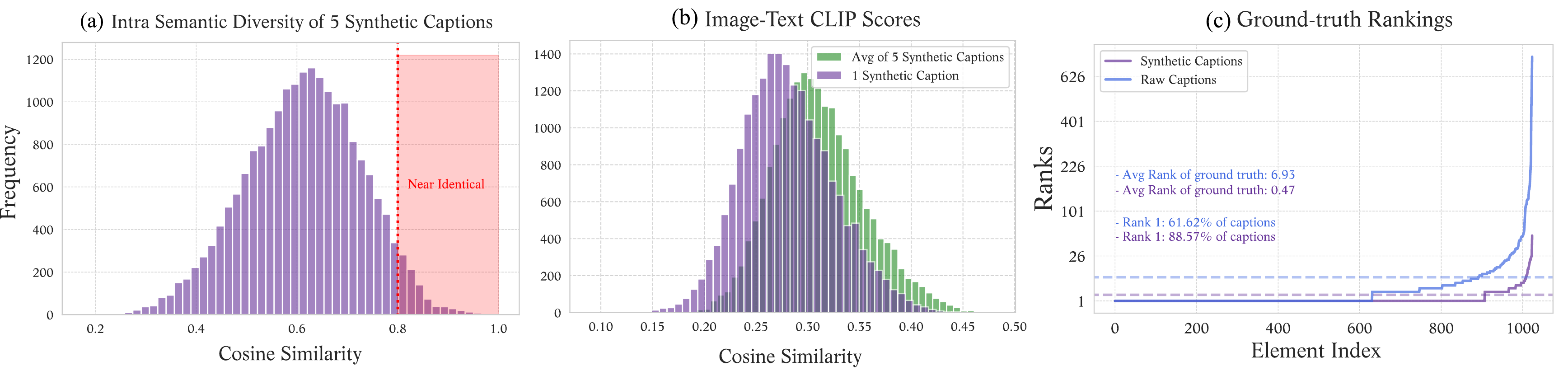}
    \caption{\textbf{Quality assessment of synthetic captions of CC3M:}
    (a) Average intra-cosine similarities between 5 synthetic captions of each image.
    (b) Cosine similarities between the features of
    each image and either the features of a single synthetic pseudo-caption or the averaged features of 5  pseudo-captions.
    In (a) and (b), we observe that using multiple synthetic positives that are diverse (a), possible
    erroneous captions can be corrected using an ensemble of  pseudo-captions that better converge to the ground truth,
    resulting in text features more aligned with their associated images (b).
    (c): The rankings of the ground-truth captions for each image in a batch of 1$k$ image-caption
    pairs. This shows that, even with relatively small batches, many negatives are well aligned with some images, and it is 
    very likely that many of these negatives are potentially correct matches for a subset of images, \ie false negatives.
    Features are computed using CLIP ViT-L/14.}
    \vspace{-0.15in}
    \label{fig:fig2}
\end{figure*}

\section{Flaws of web-collected datasets \& potential solutions}\label{sec:analysis}

Several observations drawn by analyzing the flaws of a web-collected dataset (CC3M dataset), motivating the proposed approach, are provided below:

\begin{figure}[ht!]
    \centering
    \includegraphics[width=0.5\textwidth]{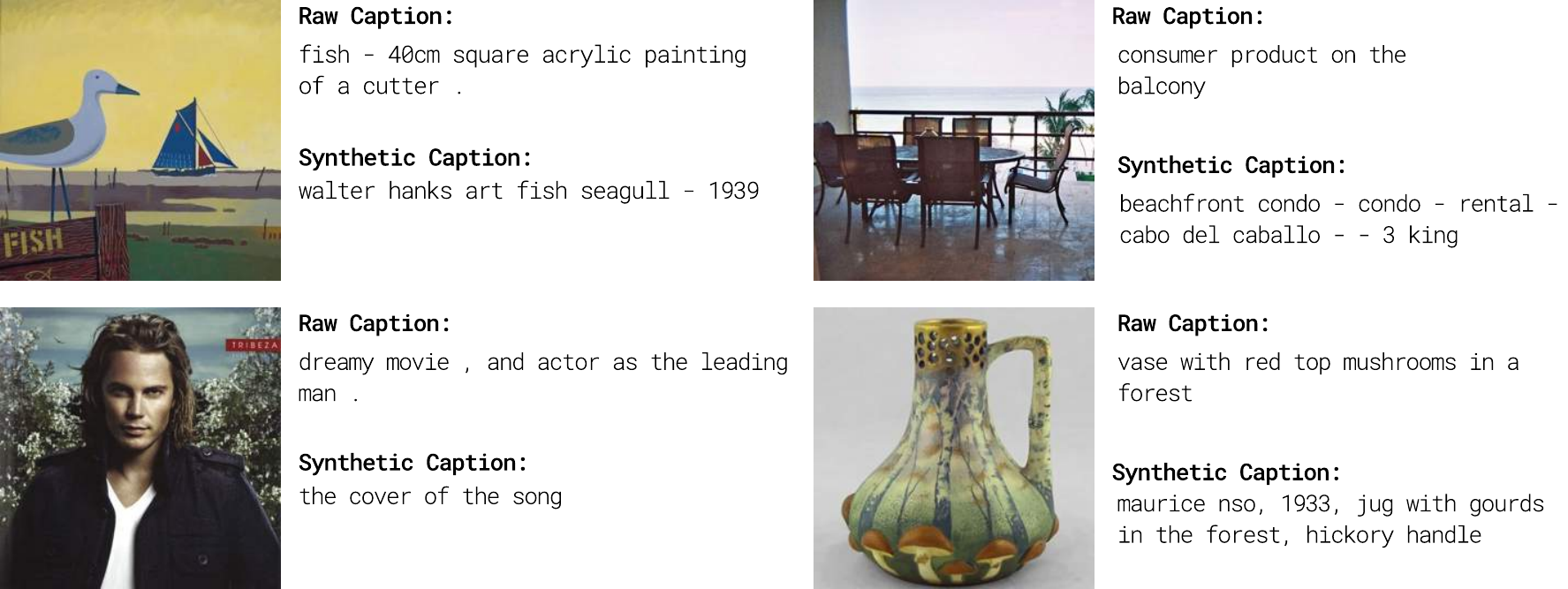}
    \caption{\textbf{Qualitative samples of synthetic captions from CC3M:}
    We show 4 examples featuring original raw and
    synthetic (BLIP2) pseudo-captions. These examples highlight typical limitations and challenges observed in synthetic captions which, while superior to raw captions, can still be considered noisy.
    }
    \vspace{-0.15in}
    \label{fig:fig3}
\end{figure}

\noindent \textbf{Original captions are noisy and repetitive:} For example, as illustrated in \cref{fig:fig1}
for the CC3M dataset, original (raw) captions contain a high number of generic captions that frequently
reoccur across the dataset (\cref{fig:fig1} (c)), and are often semantically similar (\cref{fig:fig1} (a)). Moreover, many
raw captions may be unrelated to their associated images and their visual content, as indicated by
low CLIP scores (\cref{fig:fig1} (b)).

\noindent \textbf{Re-captioning enhances quality and diversity:} A potential solution to this issue is the use of state-of-the-art image captioning models
(\eg BLIP2~\cite{li2023blip}, OFA~\cite{wang2022ofa}) to
generate synthetic pseudo-captions, which can enhance the quality and descriptiveness of the captions.
When comparing raw and pseudo-captions, it is evident that the latter are more
diverse and semantically relevant to their associated images, as shown in \cref{fig:fig1}.

\noindent \textbf{Multiple pseudo-captions should reduce noise:} State-of-the-art image captioning models, despite being capable of generating fluent and diverse captions, are often trained and
bootstrapped from the same web-collected data used in training vision-language models. Consequently, as shown in \cref{fig:fig3}, in some instances, the generated pseudo-captions 
can be ambiguous and contain hallucinations, errors, and stylistic biases similar to those found in the raw
captions. As a result, relying on a single pseudo-caption per image can still introduce a high degree of noise and can hinder the training of an effective vision-language model.

A potential solution to this issue is the use of multiple pseudo-captions or multiple positives per image in the hope that even if individual captions are incorrect, their ensemble
is of higher quality and better reflects the content of the associated image.
To probe for the possible positive effect of using multiple synthetic captions, in \cref{fig:fig2} (a),
we show the intra-cosine similarities of 5  pseudo-captions generated using beam search and, respectively, in \cref{fig:fig2} (b), the average image-text CLIP score between these synthetic captions and their associated images - contrasted with the score corresponding to a single caption.
We observe that: 1) a simple method such as beam search can generate diverse synthetic captions,
and, more crucially, 2) using multiple positives per image results in an improved ensemble
that better describes the image and helps alleviate the problem of false positives due to incorrect individual instances.

\begin{figure}[t]
    \centering
    \includegraphics[width=0.5  \textwidth]{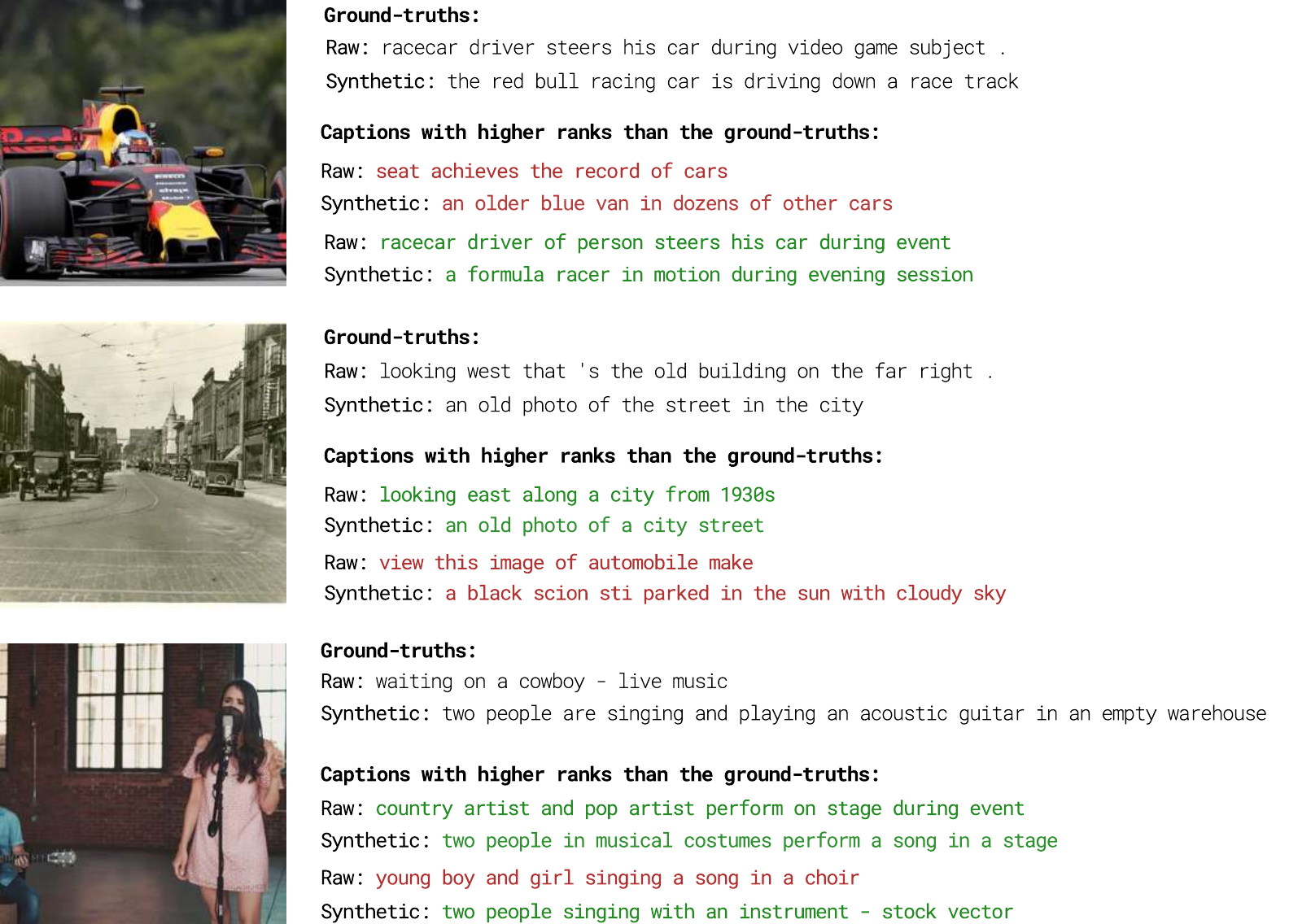}
    \caption{\textbf{Examples of high-ranking captions from CC3M:} We show 3 examples of raw and synthetic captions ranked higher than the ground-truths from a batch of 1$k$ image-caption pairs. In green, we show potential false negatives that can be used as new positives for improved training. However, possible false positives, as shown in red, can still occur. These can be handled by the robust sigmoid loss. Rankings are obtained using CLIP ViT-L/14.
    }    \label{fig:fig4}
    \vspace{-0.2in}
\end{figure}

\noindent \textbf{Mining of new positives:} As shown in \cref{fig:fig2} (c), even for a relatively small batch of 1$k$ image-caption pairs, it is common
to find captions more similar to the image than the ground-truth caption (\ie higher
ranks), and, as displayed in \cref{fig:fig4}, such high-ranking captions often
contain true positives, which are captions that can be considered ground-truth descriptions for the
associated image.

A potential solution to this is the use of online mining of new positives based on image and text feature cosine similarities. However, as shown in \cref{fig:fig4}, text-image pairs with high cosine similarity can still be false positives. To reduce them, we propose to mine the positives based on image-text, image-image, and text-text similarities, aiming to decrease the number of false negatives in the training data arising due to semantically similar images and/or captions.

%% file: sec/2_related_work.tex
\section{Related work}\label{sec:related_work}

\noindent \textbf{Contrastive pretraining under noisy web-collected data:} Current publicly available vision-language datasets are mined from the internet automatically~\cite{thomee2016yfcc100m,changpinyo2021cc12m,schuhmann2021laion,schuhmann2022laion} with only basic automatic filtering applied, which results in imperfect annotations and duplicate or near-duplicate pairs. A series of papers~\cite{andonian2022robust,wu2021data,gao2023softclip,fan2023improving} attempt to alleviate the noise present in annotations by switching from hard to soft labels, akin to knowledge distillation (KD), using various combinations of contrastive loss (i.e. InfoNCE) and KL divergence. The work in~\cite{wu2021data} constructs the soft labels using an online entropic optimal transport algorithm implemented via the Sinkhorn-Knopp algorithm. The probabilities for each image add up to 1, with 0.5 on the diagonal and the rest distributed. This assumes that, within the batch, there are always some images that are somewhat similar. In our case, we use hard labels, with multiple positives, performing reassignments only when the samples are sufficiently close, instead of forcing a distribution in all cases. Furthermore, we do not require running an optimal transport method, nor rely on a contrastive loss. The work of~\cite{andonian2022robust} progressively self-distills
soft image-text alignments to more efficiently learned robust
representations from noisy data. At every iteration, a subset of labels are ``soft'' while the rest are kept hard. Similarly, the work of~\cite{gao2023softclip} relaxes the strict one-to-one constraint, transitioning to a soft cross-modal alignment by introducing a softened target, which is generated from the
fine-grained intra-modal self-similarity. Additionally, they disentangle the negatives in the distribution to further boost the relation alignment, resulting in a combination of InfoNCE loss performed with hard labels and KL divergence. 
However, they do not perform batched text augmentations with multiple positives, as in our work, and still use a contrastive loss combined with KL, operating on soft scores. 
The works of~\cite{huynh2022boosting,chen2021incremental,hoffmann2022ranking} study the effect of removing false negatives in the context of unimodal \ie pure vision models, not considering the case of multi-modal learning. The work of~\cite{huynh2022boosting} flags (a very small number) of potential negatives using the aggregated score obtained from multiple support views per image, ~\cite{chen2021incremental} uses a clustering based approach while ~\cite{hoffmann2022ranking} is based on ranked positives, requiring a known class hierarchy (i.e. a fully supervised case) or known changes/relations (i.e. videos). The works of~\cite{huynh2022boosting,chen2021incremental} derive from the Supervised Contrastive Loss, while~\cite{hoffmann2022ranking} from InfoNCE. In contrast, our work operates on image-text data, takes into account multi-modal interactions (I2T, T2T, T2I), does not use additional support views, known hierarchies etc. and is easily scalable. 

Following a different direction, BLIP~\cite{li2022blip} and their followup~\cite{li2023blip} version, use a bootstrapping approach in which the noisy captions are filtered out using the initial model, which is then retrained on the new data. This interplay is performed offline and requires training a multitask model. The work of~\cite{santurkar2022caption} presents a small-scale study showing that random sampling of pseudo-captions improves CLIP, concluding however that scaling up the number of image-caption pairs appears to be more effective. Finally, very recently, ALIP~\cite{yang2023alip} adds a synthetic pseudo-caption and a consistency gating mechanism that weights the influence of the samples and image-text pairs on the contrastive loss. 

Different from the aforementioned methods, we propose to fix incorrectly assigned negatives and mine for new true positives using text-image, image-image, and text-text similarities. Moreover, to increase caption quality and diversity, we further propose training with multiple pseudo-captions per image within the same batch. As our methods require training with multiple positives per image, we further propose to use the sigmoid loss~\cite{zhai2023sigmoid} for training the model. 

%% file: sec/3_method.tex
\section{Method}\label{sec:method}

This section describes the proposed method, whose aim is to improve vision-language training by denoising and improving the quality of the training process/data. Specifically, Sec.~\ref{ssec:method-addressing-incorrect-pairing} addresses the problem of false negative pairs inherent to the noisy nature of large-scale image-text datasets by re-assigning them as true positives\footnote{It is possible that such cases can occur in clean datasets too, as multiple captions can describe an image and vice versa, multiple images can be described by one caption.}. Sec.~\ref{ssec:method-recaptioning} proposes \textit{text batch augmentation} for training the model with \textit{multiple} positives pairs. The effect of Secs.~\ref{ssec:method-addressing-incorrect-pairing} and~\ref{ssec:method-recaptioning} is that, for each training image, a \textit{variable} number of positive-negative pairs is formed (Sec.~\ref{ssec:method-combining}). Sec.~\ref{ssec:method-loss-function} proposes a natural way to train the model in this case by using the recently proposed sigmoid loss for vision-language pre-training. 

\begin{figure*}

     \centering
     \begin{subfigure}[t]{0.3\textwidth}
         \centering
         \includegraphics[width=6cm]{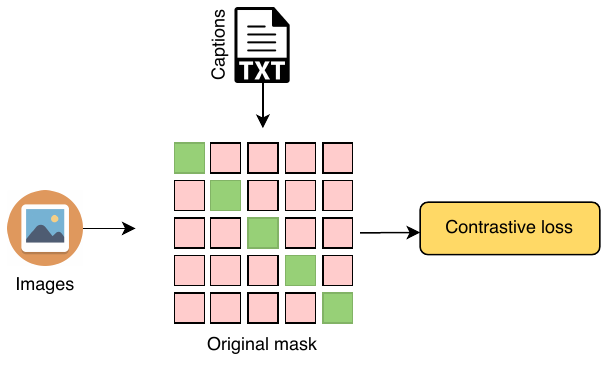}
         \caption{\textbf{Baseline~\cite{radford2021learning}}: Ground truth construction and training loss. Prior work does not take into account that some pairs may be incorrect negatives and is limited to one positive per sample.}
         \label{fig:baseline}
     \end{subfigure}
     \hfill
     \begin{subfigure}[t]{0.65\textwidth}
         \centering
         \includegraphics[width=10cm]{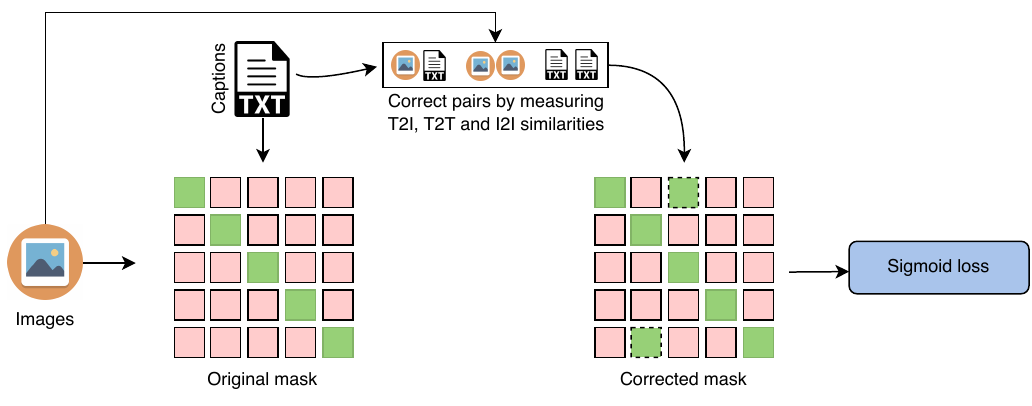}
         \caption{\textbf{Fixing incorrect negatives:} Our approach analyzes on the fly the image-text, image-image, and text-text similarities, correcting wrong negative pairs. The model is trained using the sigmoid loss (see Sec.~\ref{ssec:method-loss-function}) instead of the standard contrastive loss which is unsuitable for an arbitrary number of positive samples.}
         \label{fig:ours}
     \end{subfigure}
     ~
     \begin{subfigure}[b]{1.0\textwidth}
         \centering
         \includegraphics[width=12cm]{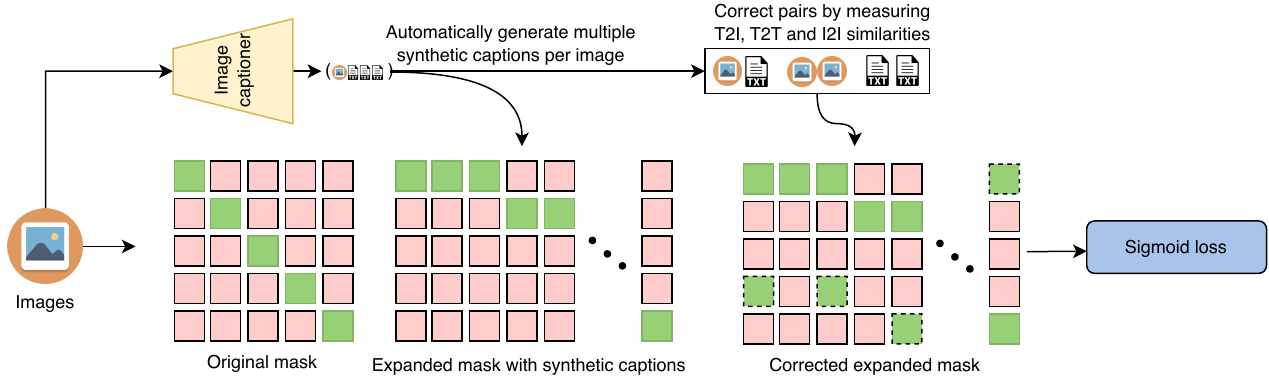}
         \caption{\textbf{Overall approach} combining fixing incorrect negatives (Sec.~\ref{ssec:method-addressing-incorrect-pairing}) with batch text argumentation (Sec.~\ref{ssec:method-recaptioning}).
         The synthetic pseudo-captions are generated offline and packed as part of the dataset.}
         \label{fig:ours_full}
         \vspace{-0.2cm}
     \end{subfigure}
        \caption{\textbf{Overview of our approach:} Fixing incorrect negatives is shown in (b). In (c) we describe our combined approach (including batch text augmentation) contrasted with the baseline of (a). Green squares denote positive pairs, while pink are negatives. Green squares with a dashed border denote identified false negatives that are corrected to true positives.}
        \label{fig:ours_overview}
        \vspace{-0.4cm}
\end{figure*}

\subsection{Fixing incorrect negatives}\label{ssec:method-addressing-incorrect-pairing}

Let $D$ be a dataset consisting of image-text pairs, with $B$ a batch of randomly selected samples $(x_i, t_i), \; i = 1, 2, \ldots, N$. In addition to the ground truth positives pairs $(x_i, t_i)$, we seek to identify and correct wrongly co-occurring negative pairs $(x_i,t_j)$ \textit{on-the-fly}. To achieve this, let us first define the image-text, image-image, and text-text cosine similarity matrices  $S_{it} = X_f \cdot T_f^T, S_{ii}=X_f \cdot X_f^T$ and $S_{tt}=T_f \cdot T_f^T$, where $S_{it}, S_{ii}, S_{tt}\in\mathbb{R}^{N \times N}$ and $X_f\in\mathbb{R}^{N \times d}$ and $T_f\in\mathbb{R}^{N \times d}$ represent the image and text features, respectively. 

Given the similarity score matrices, we define the assignment matrix $M\in \{0,1\}^{N \times N}$ as follows:
{\small
\begin{equation}
\label{eq:mask}
    M = (S_{it} > p_1) \lor (S_{ii} > p_2) \lor [(S_{tt} > p_3) \land (S_{it} > p'_1)],
\end{equation}
} where $\lor$ is the logical \texttt{or} and $\land$ the logical \texttt{and} operator, and $p_1, p'_1, p_2$ and $p_3$ are the thresholds above which a sample is marked as positive, with $p'_1 < p_1$. Note that we filter the positives found with text-text matching using image-text similarities (using threshold $p'_1$), as we observed a high portion of false positives within text-text matching, due to the fact that repeated samples often correlate with poor overall image description fidelity. The choice of $p_1, p'_1, p_2$ and $p_3$ is empirical and generally depends on the characteristics of the model. We ablate the dependency of the method on the threshold values in Sec.~\ref{sec:ablation_studies} where we show little sensitivity. Note that $M$ re-assigns a variable number of positives to each image. Fig.~\ref{fig:ours} depicts the construction process of $M$ at a high level. 

In order to calculate the cosine similarity matrices $S_{it}, S_{ii}$ and $S_{tt}$ required for the construction of $M$, we use a pre-trained model. This is akin to a form of auto-labeling/auto-filtering, where the pretrained model provides a signal for re-assessing the labeling of the samples. Although one could opt to use an EMA teacher-student approach, we found this simple approach to work sufficiently well. Moreover, some possible errors in $M$ can be handled by the robust sigmoid loss used for training (see Eq.~\ref{eq:loss_sigmoid}).

\subsection{Batch text augmentation with multiple positives} ~\label{ssec:method-recaptioning}
The currently available image-text datasets~\cite{sharma2018conceptual,changpinyo2021cc12m,schuhmann2022laion} are noisy, with high variability in the quality of the text descriptions among samples. To improve data quality, we use BLIP2~\cite{li2023blip}, an off-the-shelf image captioner, to generate multiple pseudo-captions for each image in the training set (see supplementary material for visual examples). Inspired by~\cite{hoffer2020augment}, we propose to include \textit{all pseudo-captions} as true positives \textit{within the same batch}, which we call batch text augmentation. Note that simultaneously training with multiple pseudo-captions within the same batch has not been considered in previous work. We show that this approach enables the training of highly accurate models (see Sec.~\ref{sec:results} and our ablation in Sec.~\ref{sec:ablation_studies}). In the next section, we also show how batch text augmentation can be integrated with the mask construction process defined in Sec.~\ref{ssec:method-addressing-incorrect-pairing}. Finally, we note that while batch text augmentation improves the overall performance, it does not address the presence of semantic near duplicates (i.e. false negatives) within the same training batch.

\subsection{Combined approach}\label{ssec:method-combining}

Our approach for fixing incorrect negative pairs (Sec.~\ref{ssec:method-addressing-incorrect-pairing}) and  batch text augmentation (Sec.~\ref{ssec:method-recaptioning}) can be naturally combined in order to define the total number of true positives per image. 

To this end, and without loss of generality, we assume $k$ captions per image (original caption plus pseudo-captions), hence the total number of captions and images are related by $N_{txt} = k N_{img}$. Given this, the image-text similarity matrix has now (by construction) size $S_{it} \in \mathbb{R}^{N_{img} \times N_{txt}}$. Hence, the computation of $S_{ii}$ and $S_{tt}$ needs to be adjusted to reflect this change. For the image-image case, and as $N_{img} < N_{txt}$, to make the image-image similarity matrix $S_{ii}$ have the same dimensions as $S_{it}$, i.e. $S_{ii} \in \mathbb{R}^{N_{img} \times N_{txt}}$, we replicate the scores $k$ times. In other words, a given image $x_i$ will share the score with each group of captions belonging to image $x_j$, $\forall i,j \in N_{img}$. For the text-text case, the similarity matrix is now of size $S_{tt} \in \mathbb{R}^{N_{txt} \times N_{txt}}$. Analogously, to make the $S_{tt}$ have the same dimensions as $S_{it}$, we take the average score between each caption of image $x_i$ and all $k$ captions of image $x_j$. 

Overall, we end up with similarity matrices of the same dimensions $S_{it}, S_{ii}, S_{tt}\in\mathbb{R}^{N_{img} \times N_{text}}$ and hence the assignment matrix $M$ can be again constructed by applying Eq.~\ref{eq:mask}. The overall process is depicted in Fig.~\ref{fig:ours_full}.

\subsection{Loss function}\label{ssec:method-loss-function}

The symmetrical contrastive loss (i.e. \textit{text} $\rightarrow$ \textit{image} and \textit{image} $\rightarrow$ \textit{text}) used in CLIP~\cite{radford2021learning} supports only one positive pair per sample (see Fig.~\ref{fig:baseline}), being in discordance with the requirement of training with a variable number of positive pairs per image set by the proposed methods in Secs.~\ref{ssec:method-addressing-incorrect-pairing} and~\ref{ssec:method-recaptioning}. A solution to this problem could be given by the Supervised Contrastive Loss~\cite{khosla2020supervised}, originally introduced to enable multi-view training of supervised image recognition. However, this loss is prone to noise~\cite{beyer2020we}, with the harder positive pairs dominating the signal and hindering, in part, the effect from the rest of the positive samples. This is especially problematic in the context of web-collected datasets, which are notoriously noisy. Finally, it is memory intensive and computationally demanding. In practice, we observe a 1.9$\times$ slowdown for a batch size of 8,096 samples.

A natural alternative is the BCE loss, shown to outperform cross-entropy for image classification~\cite{wightman2021resnet}, and also shown to be a viable alternative for image-text representation learning~\cite{zhai2023sigmoid}. Such formulation is particularly advantageous for the proposed approach, as the BCE loss natively supports an arbitrary number of positives per sample per batch, with the ground truth being provided simply as a binary mask. Moreover, the loss is more robust to noise in general, and hence to false negatives and positives~\cite{zhai2023sigmoid}. Finally, the initial negative bias prevents the model from being forced to learn incorrect assignments early one. Hence, we propose to use the following loss:
\begin{equation}
    \ell_{mp} = -\frac{1}{N_{txt}} \sum_{i=1}^{N_{img}} \sum_{j=1}^{N_{txt}} \log \frac{1}{1 + \text{exp} ( \; m_{ij}(-s_{ij}/\tau + \beta))},
    \label{eq:loss_sigmoid}
\end{equation}
where $m_{ij}$ is the $i,j$ element of $M$ ($-1$ for negative and $1$ for positive pairs), and respectively, $s_{ij}$ the $i,j$ element of the similarity matrix $S_{it}$.

As the negative pairs considerably outnumber the positive ones, to ensure that we start from a low initial loss (making the same observation as in~\cite{zhai2023sigmoid}), we add a learnable scalar $\beta$, set initially to a negative value. However, as the number of positive pairs is dynamic and is typically tied to both the specifics of the dataset and the threshold used to define a positive sample, different from~\cite{zhai2023sigmoid}, we propose to estimate $\beta$ at the beginning of the training process. Specifically, given the randomly initialized model, we sample $b$ batches out of the training set, and then compute and store the cosine similarities. Then, given the scores and the corresponding labels, we search for $\beta$ such that the initial loss is minimized (everything else is kept frozen). The value of $\beta$ can be found either by gradient descent or alternatively, by performing a grid search.

%% file: sec/4_results.tex
\section{Results}\label{sec:results}

\begin{table*}[ht!]
\centering
\resizebox{0.98\linewidth}{!}{
    \begin{tabular}{lccccccccccccccc}
        \toprule
        Method & \shortstack{Pre-train \\ dataset} &  \rotatebox[origin=lb]{90}{\smash{CIFAR10}} & \rotatebox[origin=lb]{90}{\smash{CIFAR100}} &  \rotatebox[origin=lb]{90}{\smash{Food101}} & \rotatebox[origin=lb]{90}{\smash{Pets}} &  \rotatebox[origin=lb]{90}{\smash{Flowers}} & 
        \rotatebox[origin=lb]{90}{\smash{SUN397}} &
        \rotatebox[origin=lb]{90}{\smash{Cars}} & 
        \rotatebox[origin=lb]{90}{\smash{DTD}} & 
        \rotatebox[origin=lb]{90}{\smash{Caltech101}} & 
        \rotatebox[origin=lb]{90}{\smash{Aircraft}} & 
        \rotatebox[origin=lb]{90}{\smash{ImageNet}}  &
        \rotatebox[origin=lb]{90}{\smash{Average}}  \\
        \midrule

        CLIP-ViT-B/32\cite{radford2021learning} & YFCC15M & 63.7 & 33.2 & 34.6 & 20.1 & 50.1 & 35.7 & 2.6 & 15.5 & 59.9 & 1.2 & 32.8 & 31.8 \\

        SLIP-ViT-B/32~\cite{mu2022slip} & YFCC15M & 50.7 & 25.5 & 33.3 & 23.5 & 49.0 & 34.7 & 2.8 & 14.4 & 59.9 & 1.7 & 34.3 & 30.0 \\
        
        FILIP-ViT-B/32~\cite{yao2021filip} & YFCC15M & 65.5 & 33.5 & 43.1 & 24.1 & 52.7 & 50.7 & 3.3 & 24.3 & 68.8 & 3.2 & 39.5 & 37.2 \\
        
        DeCLIP-ViT-B/32~\cite{li2021supervision} & YFCC15M & 66.7 & 38.7 & 52.5 & 33.8 & 60.8 & 50.3 & 3.8 & 27.7 & 74.7 & 2.1 & 43.2 & 41.3 \\
        
        DeFILIP-ViT-B/32~\cite{cui2022democratizing} & YFCC15M & 70.1 & 46.8 & 54.5 & 40.3 & \underline{63.7} & 52.4 & 4.6 & 30.2 & 75.0 & 3.3 & 45.0 & 44.2 \\
        HiCLIP-ViT-B/32~\cite{geng2023hiclip} & YFCC15M & 74.1 & 46.0 & 51.2 & 37.8 & 60.9 & 50.6 & 4.5 & 23.1 & 67.4 & 3.6 & 40.5 & 41.8 \\
        HiDeCLIP-ViT-B/32~\cite{geng2023hiclip} & YFCC15M & 65.1 & 39.4 & \underline{56.3} & \underline{43.6} & \textbf{64.1} & \underline{55.4} & \underline{5.4} & \textbf{34.0} & \underline{77.0} & \textbf{4.6} & \underline{45.9} & \underline{44.6} \\
        ALIP-ViT-B/32~\cite{yang2023alip} & YFCC15M & \textbf{83.8} & \underline{51.9} & 45.4 & 30.7 & 54.8 & 47.8 & 3.4 & 23.2 & 74.1 & 2.7 & 40.3 & 41.7  \\
        \midrule
        FFF-ViT-B/32 (Ours) & YFCC15M & \underline{75.8} & \textbf{56.3} & \textbf{58.6} & \textbf{59.8} & 62.1 & \textbf{61.5} & \textbf{16.3} & \underline{33.4} &  \textbf{79.6} & \textbf{4.6} & \textbf{51.1} & \textbf{50.8}\\
        \bottomrule
  \end{tabular}} 
  \caption{Zero-shot classification performance on 11 downstream datasets. All models were pre-trained on YFCC15M.}
\label{tab:zero-shot-sota-yfcc}
\end{table*}

\begin{table*}[t]
    \centering
    \resizebox{0.98\linewidth}{!}{
        \centering
        \begin{tabular}{lcccccccccccc}
        \toprule
        & \multicolumn{6}{c}{Text retrieval} & \multicolumn{6}{c}{Image retrieval} \\
        & \multicolumn{3}{c}{Flickr30k} & \multicolumn{3}{c}{MSCOCO} & \multicolumn{3}{c}{Flickr30k} & \multicolumn{3}{c}{MSCOCO} \\
        Method &  R@1 & R@5 & R@10 & R@1 & R@5 & R@10 & R@1 & R@5 & R@10 & R@1 & R@5 & R@10 \\
        \midrule
        CLIP-ViT-B/32\cite{radford2021learning} & 34.9 & 63.9 & 75.9 & 20.8 & 43.9 & 55.7 & 23.4 & 47.2 & 58.9 & 13.0 & 31.7 & 42.7 \\
        SLIP-ViT-B/32~\cite{mu2022slip} & 47.8 & 76.5 & 85.9 & 27.7 & 52.6 & 63.9 & 32.3 & 58.7 & 68.8 & 18.2 & 39.2 & 51.0 \\
        DeCLIP-ViT-B/32~\cite{li2021supervision} & 51.4 & 80.2 & 88.9 & 28.3 & 53.2 & 64.5 & 34.3 & 60.3 & 70.7 & 18.4 & 39.6 & 51.4 \\
        UniCLIP-ViT-B/32~\cite{lee2022uniclip} & 52.3 & 81.6 & 89.0 & 32.0 & 57.7 & 69.2 & 34.8 & 62.0 & 72.0 & 20.2 & 43.2 & 54.4 \\
        HiCLIP-ViT-B/32~\cite{geng2023hiclip} & - & - & - & 34.2 & 60.3 & 70.9 & - & - & - & 20.6 & 43.8 & 55.3 \\
        HiDeCLIP-ViT-B/32~\cite{geng2023hiclip} & - & - & - & 38.7 & 64.4 & 74.8 & - & - & - & 23.9 & 48.2 & 60.1 \\
        ALIP-ViT-B/32~\cite{yang2023alip} & 70.5 & 91.9 & 95.7 & 46.8 & 72.4 & 81.8 & 48.9 & 75.1 & 82.9 & 29.3 & 54.4 & 65.4 \\
        \midrule
        FFF-ViT-B/32  (Ours) & \textbf{85.3} & \textbf{97.5} & \textbf{99.4} & \textbf{61.7} & \textbf{84.5} & \textbf{90.4} & \textbf{67.6} & \textbf{89.1} & \textbf{93.3} & \textbf{44.3} & \textbf{70.9} &\textbf{80.1} \\
        \bottomrule
        \end{tabular}}
    \caption{Zero-shot image-text retrieval on the test splits of Flickr30k and MSCOCO. All models were pre-trained on YFCC15M.}
    \label{tab:retrieval}
    \vspace{-0.5cm}
\end{table*}


\noindent \textbf{Pretraining Datasets:} To allow for fair comparisons with prior work, we pre-train our approach on YFCC15M-v2~\cite{li2021align}, a subset of YFCC100M~\cite{thomee2016yfcc100m} containing approximately 15M image-text pairs. To cover different dataset sizes, we also conduct experiments on CC3M~\cite{sharma2018conceptual} and CC12M~\cite{changpinyo2021cc12m}, and in the supplementary material, on Open30M and Open70M datasets, further showcasing our method's scalability with respect to the dataset size.

\noindent \textbf{Implementation details:} Architecturally, we use the same model topology and setting as in CLIP~\cite{radford2021learning}, specifically, using AdamW~\cite{loshchilov2018fixing}, learning rate of $1e-3$ and weight decay of $0.1$, except for CC3M where we set the weight decay to $0.5$, as in prior work~\cite{li2021supervision}. In terms of augmentations, we follow~\cite{li2021supervision}, randomly resizing and cropping the image to $224\times224$px, applying random flipping, random Gaussian blur (between 0.1 and 2.0) and color jittering (0.4, 0.4, 0.4, 0.1). For text, the data is truncated to 77 tokens. Note, that the branch used to construct the assignment matrix $M$ uses no augmentations (i.e. resize to $256\times256$px, followed by center crop, resulting in a $224\times 224$px image). The thresholds were set to $p_1 = 0.27$, $p_2=0.92$, $p_3=0.99$, $p'_1=0.24$. Unless otherwise specified, the models are trained for 32 epochs with a batch size of 8,096 on 8 NVIDIA A100 GPUs. All of our models and training code are implemented using PyTorch~\cite{paszke2017automatic}.

\subsection{Comparison with state-of-the-art}

Following recent work on vision-language pretraining~\cite{mu2022slip,yang2023alip,geng2023hiclip}, we compare our method with state-of-the-art approaches for zero-shot classification and zero-shot retrieval. See supplementary material for linear probe evaluation.

\noindent \textbf{Zero-shot classification:} For zero-shot classification evaluation, for the main setting, we select the common subset of datasets that facilitate a direct comparison with prior state-of-the-art. In particular, we evaluate our approach on the following datasets: CIFAR-10~\cite{krizhevsky2009learning}, CIFAR-100~\cite{krizhevsky2009learning}, Food101~\cite{bossard2014food}, Pets~\cite{parkhi2012cats}, Flowers~\cite{nilsback2008automated}, SUN397~\cite{xiao2010sun}, Stanford Cars~\cite{krause20133d}, DTD~\cite{cimpoi2014describing}, Caltech101~\cite{fei2004learning}, FGVC-Aircraft~\cite{maji2013fine} and ImageNet~\cite{deng2009imagenet}. The evaluation is performed using the same prompt templates and class names as in prior work~\cite{yang2023alip,geng2023hiclip,mu2022slip}. 

As the results from Tab.~\ref{tab:zero-shot-sota-yfcc} show, our approach outperforms all prior methods, improving by $6.2\%$ in absolute terms on top of the previous best result of HiDeCLIP~\cite{geng2023hiclip} (which benefits from a better architecture) when aggregated across 11 datasets. Notably, we set a new state-of-the-art result on ImageNet, too ($51.1\%$). 
Finally, we significantly improve upon ALIP~\cite{yang2023alip}, which also makes use of synthetic captions, outperforming it by $9.1\%$.

\begin{table}[ht!]
\vspace{-0.3cm}
\centering
\begin{tabular}{l|cc}
\toprule
Method &  CC3M & CC12M \\
\midrule
CLIP~\cite{radford2021learning} & 20.6 & 36.5 \\
ProtoCLIP~\cite{chen2022prototypical} & 21.5 & - \\
CyCLIP~\cite{goel2022cyclip} & 22.1 & - \\
CLOOB~\cite{furst2022cloob} & 24.0 & -\\
SoftCLIP~\cite{gao2023softclip} & 24.2 & 43.2\\
DeCLIP~\cite{li2021supervision} & 27.2 & 41.0 \\
CLIP-Rocket~\cite{fini2023improved} & 27.4 & 44.4 \\
BoW~\cite{tejankar2021fistful} & 30.3 & - \\
FFF (Ours) & \textbf{33.4} & \textbf{47.4} \\
\bottomrule
\end{tabular}
\caption{Zero-shot evaluation on Imagenet in terms of Top-1 (\%) accuracy for a ResNet-50 model pre-trained on CC3M/CC12M.}
\label{tab:sota-cc3m-cc12m}
\vspace{-0.3cm}
\end{table}

For completeness, we also adhere to the protocol of pretraining a ResNet-50 on CC3M, and respectively, CC12M and then evaluating it for zero-shot classification on ImageNet. As the results from Tab.~\ref{tab:sota-cc3m-cc12m} show, the same conclusions hold. Our method outperforms the previous best result by 3.1\% on CC3M (30.3\% vs 33.4\%) and 3.0\% on CC12M (44.4\% vs 47.4\%). See supplementary material for results on Open30M and Open70M.

\noindent \textbf{Zero-shot retrieval:} Consistent with prior work, we evaluate our approach for zero-shot retrieval on Flickr-30k~\cite{young2014image} and MS-COCO~\cite{lin2014microsoft} reporting results in terms of R@\{1,5,10\} for both text and image retrieval. The results are summarized in Tab.~\ref{tab:retrieval}. As it can be observed, our approach offers significant gains across all metrics and datasets used, improving on top of the prior state-of-the-art ALIP~\cite{yang2023alip} by 14.8\% and 18.7\% in terms of R@1 on Flickr30k for text, and respectively, image retrieval. Similarly, we outperform the previous best result by 14.9\% and 15.0\% in terms of R@1 on MSCOCO for text and image retrieval. This highlights that our approach results in representations that can capture subtle and fine-grained details.

%% file: sec/5_ablations_and_conclusion.tex
\section{Ablation studies}\label{sec:ablation_studies}

For our ablation studies, the results reported are produced using a ViT-B/16 model pretrained on CC3M dataset.

\noindent \textbf{Effect of fixing incorrect negatives:} herein, we analyze the effectiveness of the proposed algorithm of Sec.~\ref{ssec:method-addressing-incorrect-pairing}. By analyzing the result from Tab.~\ref{tab:ablation-cc3m-gt-corrections}, we can observe consistent gains for all 3 cases of interest: a) when using the web-collected captions (+2.7\% gain), b) when using one pseudo-caption (+3.5\% improvement) and c) when all available pseudo-captions at once (+1.8\%). Overall, compared to the baseline accuracy of 18.6\%, our approach improves by +14.3\% (top-1 accuracy of 32.9\%). The results show that our approach provides gains across all options considered. 

\begin{table}[h]
\vspace{-0.3cm}
\centering
\begin{tabular}{cc|c}
\toprule
Fix incorrect negatives & Num. captions & Top-1 (\%) \\
\midrule
\xmark & 0 & 18.6\\
\checkmark & 0 & 21.3\\
\xmark & 1 & 23.3\\
\checkmark & 1 & 26.8 \\
\xmark & 5 & 31.1\\
\checkmark & 5 & \textbf{32.9}\\
\bottomrule
\end{tabular}
\caption{\textbf{Effect of fixing incorrect negatives:} Zero-shot evaluation on ImageNet in terms of Top-1 (\%) accuracy.}
\label{tab:ablation-cc3m-gt-corrections}
\vspace{-0.2cm}
\end{table}

\noindent \textbf{Effect of different components in Eq.~\ref{eq:mask}:} In Eq.~\ref{eq:mask}, the constructed assignment matrix $M$ is computed from three feature similarity  matrices $S_{it}$, $S_{ii}$ and $S_{tt}$. Herein, we evaluate the impact of each of these components. As the results from Tab.~\ref{tab:ablation-mask} show, viewed independently, the $S_{it}$ is the most impactful, as it has a dual effect, both in terms of filtering incorrect pairs and of adjusting for semantically similar samples. Moreover, the results hold for both ground truth captions and pseudo-captions. 

\begin{table}[h]
\centering
\begin{tabular}{cc|c}
\toprule
Assign. Matrix $M$ & Num. captions & Top-1 (\%) \\
\midrule
None & 0 & 18.6\\
$S_{tt} > p_3$ & 0 & 18.8\\
$S_{ii} > p_2$ & 0 & 21.3\\
$S_{it} > p_1$ & 0 & 21.4 \\
Eq.~\ref{eq:mask} (all) & 0 & 22.0\\
\midrule
None & 1 & 23.1\\
$S_{tt} > p_3$ & 1 & 23.6\\
$S_{ii} > p_2$ & 1 & 24.6\\
$S_{it} > p_1$ & 1 & 26.0 \\
Eq.~\ref{eq:mask} (all) & 1 & 26.8\\
\bottomrule
\end{tabular}
\caption{\textbf{Effect of different components in Eq.~\ref{eq:mask}:} Zero-shot evaluation on ImageNet in terms of Top-1 (\%) accuracy.}
\label{tab:ablation-mask}
\end{table}

\noindent \textbf{Effect of batch text augmentation:} Herein, we assess the impact of training with multiple pseudo-captions within the same batch, as described in Sec.~\ref{ssec:method-recaptioning}. Tab.~\ref{tab:ablation-cc3m-num-captions} shows accuracy vs number of pseudo-captions used during training. As we can observe, increasing the number of captions increases the accuracy of the model, inline with the expectations. 

As an additional baseline, we compare against a model trained by randomly sampling 1 out of 5 captions (as opposed to using them jointly as proposed in our work) on CC3M and YFCC-15M. On CC3M the performance drops by 1.5\%, from 32.9\% to 31.4\%, while on YFCC-v2 from 51.1\% to 44.1\%. This further highlights the importance of the proposed batch text augmentation. 

\noindent \textbf{Effect of image captioner:} We also compare the effect of using two different state-of-the-art image captioners, OFA~\cite{wang2022ofa} and BLIP-2~\cite{li2023blip}. As the results from Tab.~\ref{tab:ablation-ofa-vs-blip} show, both captioners lead to identical performance.

\begin{table}[h]
\centering
\begin{tabular}{c|cccc}
\toprule
 Num. captions & 0 & 1 & 3 & 5 \\
\midrule
Top-1 (\%) & 18.6 & 23.3 & 30.2 & 31.1 \\
\bottomrule
\end{tabular}
\caption{\textbf{Effect of batch text augmentation:} Zero-shot evaluation on ImageNet in terms of Top-1 (\%) accuracy.}
\label{tab:ablation-cc3m-num-captions}
\vspace{-0.5cm}
\end{table}

\begin{table}[h]
\centering
\begin{tabular}{c|c}
\toprule
Image captioner & Top-1 (\%) \\
\midrule
OFA~\cite{wang2022ofa} & 32.9\\
BLIP-2~\cite{li2023blip} & 32.9 \\
\bottomrule
\end{tabular}
\caption{\textbf{Effect of the image captioner:} Zero-shot evaluation on ImageNet in terms of Top-1 (\%) accuracy.}
\label{tab:ablation-ofa-vs-blip}
\vspace{-0.2cm}
\end{table}

\noindent \textbf{Comparison with the supervised contrastive loss:} To further validate the loss choice, we compare against a model trained with the supervised contrastive loss~\cite{khosla2020supervised}. For a fair comparison, both models were trained using the same settings on CC3M. When evaluated for zero-shot classification on ImageNet, the supervised contrastive model achieved only 19.0\% accuracy vs  21.3\% achieved by our model. Note, that similar results are obtained using a InfoNCE based loss. This result empirically solidifies the arguments made in Sec.~\ref{ssec:method-loss-function}.

\section{Conclusions}\label{sec:conclusions}

In this work, we propose a new approach to vision-language pretraining based on multi-positive sample pairing that fixes incorrect negatives and addresses low caption quality. The latter is tackled by a newly introduced batch text augmentation strategy, in which multiple new positive pairs are concomitantly added via synthetic recaptioning. Departing from the typical contrastive loss, to enable efficient training under an arbitrary number of positives per sample, we propose to train the model with a sigmoid loss. In the process, we highlight the crucial role of noise and caption quality in vision-language pre-training, offering an in-depth  analysis. All in all, we show large improvements over the current state-of-the-art method for both zero-shot image recognition ($\sim +6\%$ on average of 11 datasets) and  retrieval ($\sim +19\%$ on Flickr30k and $\sim +15\%$ on MSCOCO).

%% file: sec/X_suppl.tex
\appendix

\section{Additional comparisons with state-of-the-art}

\subsection{Zero-shot recognition on Open30M and Open70M datasets}

To further showcase the scalability of our approach, we follow~\cite{lee2022uniclip,geng2023hiclip}, pretraining our method on a combination of 4 publicly available datasets, dubbed Open30M (see Tab.~\ref{tab:datasets} for composition). The pretraining hyperparameters remain the same as for YFCC. Once trained, we evaluate it in a zero-shot manner on the same suite of 11 datasets. As the results from Tab.~\ref{tab:zero-shot-sota-open30m} show, our approach outperforms all prior methods, improving upon the prior best result of~\cite{geng2023hiclip} by +4.7\% aggregated over 11 datasets, including by +3.1\% on ImageNet.

Finally, we extend the Open30M images dataset by adding RedCaps~\cite{desai2021redcaps}, OpenImages-8M~\cite{kuznetsova2020open} and YFCC-v1, creating Open70M. As the results from Tabs.~\ref{tab:zero-shot-sota-open30m} and~\ref{tab:retrieval-all} show, our approach scales well, with consistent gains for both zero-shot retrieval and classification.

\begin{table*}[h!]
\centering
\resizebox{0.98\linewidth}{!}{
    \begin{tabular}{lccccccccccccccc}
        \toprule
        Method & \shortstack{Pre-train \\ dataset} &  \rotatebox[origin=lb]{90}{\smash{CIFAR10}} & \rotatebox[origin=lb]{90}{\smash{CIFAR100}} &  \rotatebox[origin=lb]{90}{\smash{Food101}} & \rotatebox[origin=lb]{90}{\smash{Pets}} &  \rotatebox[origin=lb]{90}{\smash{Flowers}} & 
        \rotatebox[origin=lb]{90}{\smash{SUN397}} &
        \rotatebox[origin=lb]{90}{\smash{Cars}} & 
        \rotatebox[origin=lb]{90}{\smash{DTD}} & 
        \rotatebox[origin=lb]{90}{\smash{Caltech101}} & 
        \rotatebox[origin=lb]{90}{\smash{Aircraft}} & 
        \rotatebox[origin=lb]{90}{\smash{ImageNet}}  &
        \rotatebox[origin=lb]{90}{\smash{Average}}  \\
        \midrule

        CLIP-ViT-B/32~\cite{radford2021learning} & Open30M & 77.3 & 48.1 & 59.1 & 58.5 & 58.2 & 52.6 & 17.7 & 28.0 & 80.8 & 3.2 & 48.8 & 48.4 \\
        HiCLIP-ViT-B/32~\cite{geng2023hiclip} & Open30M & 77.6 & 56.2 & 63.9 & 65.6 & 62.5 & 60.7 & 22.2 & 38.0 & 82.4 & 5.5 & 52.9 & 53.4 \\
        UniCLIP-ViT-B/32~\cite{lee2022uniclip} & Open30M & 87.8 & 56.5 & 64.6 & 69.2 & 8.0 & 61.1 & 19.5 & 36.6 & 84.0 & 4.7 & 54.2 & 49.7 \\
        HiDeCLIP-ViT-B/32~\cite{geng2023hiclip} & Open30M & 80.4 & 54.2 & 68.9 & 73.5 & \textbf{66.1} & 65.2 & 26.8 & 44.1 & \textbf{87.8} & \textbf{7.2} & 56.9 & 57.4 \\
           \midrule
        FFF-ViT-B/32 (Ours) & Open30M & \textbf{92.4} & \textbf{73.6} & \textbf{70.4} & \textbf{79.9} & 64.1 & \textbf{67.7} & \textbf{41.2} & \textbf{44.3} &  84.1 & 5.2 & \textbf{60.0} & \textbf{62.1} \\
        \midrule
        FFF-ViT-B/32 (Ours) & Open70M & \textbf{92.7} & \textbf{73.7} & \textbf{79.8} & \textbf{78.8} & \textbf{68.3} & \textbf{68.7} & \textbf{47.3} & \textbf{51.1} &  \textbf{86.5} & \textbf{5.3} & \textbf{65.9} & \textbf{65.3} \\
        \bottomrule
  \end{tabular}} 
  \caption{Zero-shot classification performance on 11 downstream datasets. Results taken from~\cite{geng2023hiclip}.}
\label{tab:zero-shot-sota-open30m}
\end{table*}

\begin{table*}[ht!]
\centering
\resizebox{0.98\linewidth}{!}{
    \begin{tabular}{lcccccccccccc}
        \toprule
        Method & \shortstack{Pre-train \\ dataset} &  \rotatebox[origin=lb]{90}{\smash{CIFAR10}} & \rotatebox[origin=lb]{90}{\smash{CIFAR100}} &  \rotatebox[origin=lb]{90}{\smash{Food101}} & \rotatebox[origin=lb]{90}{\smash{Pets}} &  \rotatebox[origin=lb]{90}{\smash{Flowers}} & 
        \rotatebox[origin=lb]{90}{\smash{SUN397}} &
        \rotatebox[origin=lb]{90}{\smash{Cars}} & 
        \rotatebox[origin=lb]{90}{\smash{DTD}} & 
        \rotatebox[origin=lb]{90}{\smash{Caltech101}} & 
        \rotatebox[origin=lb]{90}{\smash{Aircraft}} & 
        \rotatebox[origin=lb]{90}{\smash{Average}}  \\
        \midrule
        CLIP-ViT-B/32~\cite{radford2021learning} & YFCC15M & 86.5 & 64.7 & 69.2 & 64.6 & 90.6 & 66.0 & 24.9 & 61.3 & 79.1 & 23.1 & 63.0 \\
        DeCLIP-ViT-B/32~\cite{li2021supervision} & YFCC15M & 89.2 & 69.0 & 75.4 & 72.2 & 94.4 & 71.6 & 31.0 & 68.8 & 87.9 & 27.6 & 68.7 \\
        HiCLIP-ViT-B/32~\cite{geng2023hiclip} & YFCC15M & 89.5 & 71.1 & 73.5 & 70.6 & 91.9 & 68.8 & 30.8 & 63.9 & 84.8 & 27.4 & 67.2 \\
        HiDeCLIP-ViT-B/32~\cite{geng2023hiclip} & YFCC15M & 88.1 & 70.7 & 77.6 & 75.5 & \textbf{95.6} & 72.2 & 36.0 & 70.1 & 90.0 & 32.6 & 70.8 \\
        ALIP-ViT-B/32~\cite{yang2023alip} & YFCC15M & \textbf{94.3} & 77.8 & 75.8 & 76.0 & 95.1 & 73.3 & 33.6 & 71.7 & 88.5 & 36.1 & 72.2  \\
        \midrule
        FFF-ViT-B/32 (Ours) & YFCC15M & 93.9 & \textbf{78.4} & \textbf{80.3} & \textbf{84.9} & 94.7 & \textbf{96.2} & \textbf{55.5} & \textbf{72.2} & \textbf{99.9} & \textbf{36.5} & \textbf{79.2} \\
        \bottomrule
  \end{tabular}} 
  \caption{Linear probe classification performance on various downstream datasets. All models were pre-trained on YFCC15M. Results taken from~\cite{yang2023alip}.}
\label{tab:linear-probe-yfcc}
\end{table*}

\begin{table*}[t]
    \centering
    \resizebox{0.98\linewidth}{!}{
        \centering
        \begin{tabular}{lccccccccccccc}
        \toprule
        & & \multicolumn{6}{c}{Text retrieval} & \multicolumn{6}{c}{Image retrieval} \\
        & & \multicolumn{3}{c}{Flickr30k} & \multicolumn{3}{c}{MSCOCO} & \multicolumn{3}{c}{Flickr30k} & \multicolumn{3}{c}{MSCOCO} \\
        Method & \shortstack{Pre-train \\ dataset}   & R@1 & R@5 & R@10 & R@1 & R@5 & R@10 & R@1 & R@5 & R@10 & R@1 & R@5 & R@10 \\
        \midrule
        FFF-ViT-B/32  (Ours) & YFCC-15M &\textbf{85.3} & \textbf{97.5} & \textbf{99.4} & \textbf{61.7} & \textbf{84.5} & \textbf{90.4} & \textbf{67.6} & \textbf{89.1} & \textbf{93.3} & \textbf{44.3} & \textbf{70.9} &\textbf{80.1} \\
          FFF-ViT-B/32  (Ours) & Open30M &\textbf{87.9} & \textbf{99.2} & \textbf{99.6} & \textbf{64.2} & \textbf{85.8} & \textbf{91.7} & \textbf{72.0} & \textbf{91.4} & \textbf{94.9} & \textbf{46.4} & \textbf{72.6} &\textbf{81.6} \\
            FFF-ViT-B/32  (Ours) & Open70M &\textbf{87.5} & \textbf{98.1} & \textbf{99.3} & \textbf{66.6} & \textbf{86.6} & \textbf{91.6} & \textbf{72.9} & \textbf{92.4} & \textbf{95.7} & \textbf{49.1} & \textbf{74.9} &\textbf{83.2} \\
        \bottomrule
        \end{tabular}}
    \caption{Zero-shot image-text retrieval on the test splits of Flickr30k and MSCOCO for models pretrained on YFCC-15M, Open30M and Open70M.}
    \label{tab:retrieval-all}
\end{table*}

\begin{table*}[ht!]
\centering
\resizebox{0.98\linewidth}{!}{
    \begin{tabular}{lcccccccccccc}
        \toprule
        Method & \shortstack{Pre-train \\ dataset} &  \rotatebox[origin=lb]{90}{\smash{CIFAR10}} & \rotatebox[origin=lb]{90}{\smash{CIFAR100}} &  \rotatebox[origin=lb]{90}{\smash{Food101}} & \rotatebox[origin=lb]{90}{\smash{Pets}} &  \rotatebox[origin=lb]{90}{\smash{Flowers}} & 
        \rotatebox[origin=lb]{90}{\smash{SUN397}} &
        \rotatebox[origin=lb]{90}{\smash{Cars}} & 
        \rotatebox[origin=lb]{90}{\smash{DTD}} & 
        \rotatebox[origin=lb]{90}{\smash{Caltech101}} & 
        \rotatebox[origin=lb]{90}{\smash{Aircraft}} & 
        \rotatebox[origin=lb]{90}{\smash{Average}}  \\
       \midrule
        CLIP-ViT-B/32~\cite{radford2021learning} & Open30M & 92.0 & 74.7 & 78.8 & 80.7 & 93.7 & 72.6 & 55.9 & 71.4 & 88.6 & 29.7 & 73.8 \\
        HiCLIP-ViT-B/32~\cite{geng2023hiclip} & Open30M & 92.8 & 75.8 & 80.5 & 81.3 & 94.4 & 73.6 & 59.4 & 72.2 & 90.3 & 33.6 & 75.4 \\
        DeCLIP-ViT-B/32~\cite{li2021supervision} & Open30M & 93.1 & 76.9 & 82.0 & 82.7 & 96.0 & 74.9 & 59.8 & 74.5 & 92.6 & 32.7 & 76.5 \\
        HiDeCLIP-ViT-B/32~\cite{geng2023hiclip} & Open30M & 92.7 & 75.6 & 82.9 & 83.3 & 95.7 & 75.6 & 62.8 & 74.5 & 92.0 & 35.8 & 77.1 \\
        \midrule
        FFF-ViT-B/32 (Ours) & Open30M & \textbf{96.6} & \textbf{84.1} & \textbf{83.8} & \textbf{87.4} & \textbf{95.7} & \textbf{97.3} & \textbf{74.1} & \textbf{75.5} & \textbf{99.9} & \textbf{38.7} & \textbf{83.3} \\

        \bottomrule
  \end{tabular}} 
  \caption{Linear probe classification performance on various downstream datasets. All models were pre-trained on Open30M. Results taken from~\cite{geng2023hiclip}.}
\label{tab:linear-probe-open30m}
\end{table*}

\subsection{Linear probe}

In addition to zero-shot evaluation, we also present  linear
probe results in \cref{tab:linear-probe-yfcc} for models pre-trained on
YFCC15M and in \cref{tab:linear-probe-open30m} for models pre-trained on
Open30M. Similar to zero-shot experiments, we use the \texttt{clip-benchmark}
repository\footnote{\url{https://github.com/LAION-AI/CLIP_benchmark}} to run these experiments.
For each dataset, we cache the features of the training and test sets,
and then use the training set's features and its ground-truth labels to train
a linear layer on top. The linear linear is trained for
20 epochs using the standard cross-entropy loss
and AdamW optimizer with a learning rate of 0.1, no weight decay, and a cosine learning rate scheduler. The trained linear layer is then
used over the cached test features to obtain the accuracy.
Similar to zero-shot experiments, our approach
outperforms previous methods by large margins, \ie,
+$7.0\%$ with YFCC15M pertaining (\cref{tab:linear-probe-yfcc})
and +$6.2\%$ with Open30M pertaining over 11
image classification datasets.

\section{Additional ablation studies}

\noindent \textbf{Sensitivity to the threshold value:} The selection of threshold values is intuitive, and the model is generally forgiving within a certain plateau of values. For $S_{tt}$ and $S_{ii}$, they are simply set to high values to target nearly identical samples. For $S_{it}$, we start from the mean score of the positive pairs, which is $0.29$, and explore a few adjacent values, noting that all values located in the same vicinity perform well as shown in \cref{tab:ablation-cc3m-threshold}.

\begin{table}[h]
\centering
\begin{tabular}{ccccc}
\toprule
 0.26 & 0.27 & 0.28 & 0.29 & 0.3 \\
\midrule
32.4 & 32.9 & 32.8 & 32.8 & 32.5 \\
\bottomrule
\end{tabular}
\caption{\textbf{Effect of the $S_{it}$ threshold ($p_1$):} Zero-shot evaluation on Imagenet in terms of Top-1 (\%) accuracy.} 
\label{tab:ablation-cc3m-threshold}
\vspace{-0.4cm}
\end{table}

\section{Zero-shot classification prompts}

For zero-shot recognition, we align with prior work~\cite{radford2021learning,yang2023alip}, using the same list of prompts. The full list is defined in Tab.~\ref{tab:prompt}.

\section{Zero-shot retrieval evaluation considerations}

As the synthetic captions are generated by models pretrained on external data, a reasonable question to ask is wherever there is potential data leakage. For the Flick30k dataset, no such issues are present, as BLIP2 did not use any data from the training set of Flickr30k during any of its training phases. For MSCOCO, we note that only 100k out of 120M samples used for training BLIP2 were images from the COCO training set, hence the impact is likely minimal, if any. We note here that the current state-of-the-art method, ALIP, is subject to the same potential issue, as they also make use of synthetic captions produced by a model that was pre-trained on MSCOCO data (i.e. OFA).

\begin{table}[ht!]
\centering
\resizebox{0.7\linewidth}{!}{
    \begin{tabular}{lr}
    \toprule
    Pre-train dataset & Number of examples \\
    \midrule
    SBU & 844,574 \\
    CC12M & 10,503,723 \\
    CC3M & 2,876,999 \\
    YFCC15M-V2 & 14,864,773 \\
    \midrule
    Open30M & 29,090,069 \\
    \bottomrule
  \end{tabular}} 
  \caption{Number of examples per each training dataset. Open30M
  is the combination of all four datasets, \ie, SBU, CC3M, CC12M and YFCC15M-V2.}
\label{tab:datasets}
\end{table}

\begin{table*}
\centering
\resizebox{0.9\linewidth}{!}{
\begin{tabular}{llll}
\toprule
\multicolumn{4}{l}{\bf CIFAR 10 \& CIFAR 100} \\
a photo of a \{label\}. &
a blurry photo of a \{label\}. &
a black and white photo of a \{label\}. &
a low contrast photo of a \{label\}. \\
a high contrast photo of a \{label\}. &
a bad photo of a \{label\}. &
a good photo of a \{label\}. &
a photo of a small \{label\}. \\
a photo of a big \{label\}.&
a photo of the \{label\}.&
a blurry photo of the \{label\}.&
a black and white photo of the \{label\}. \\
a low contrast photo of the \{label\}.&
a high contrast photo of the \{label\}.&
a bad photo of the \{label\}.&
a good photo of the \{label\}. \\
a photo of the small \{label\}.&
a photo of the big \{label\}.& & \\
\midrule
\multicolumn{4}{l}{\bf Food101} \\
a photo of \{label\}, a type of food. & & \\
\midrule
\multicolumn{4}{l}{\bf Caltech101} \\
a photo of a \{label\}. &
a painting of a \{label\}. &
a plastic \{label\}. &
a sculpture of a \{label\}. \\
a sketch of a \{label\}. &
a tattoo of a \{label\}. &
a toy \{label\}. &
a rendition of a \{label\}. \\
a embroidered \{label\}. &
a cartoon \{label\}. &
a \{label\} in a video game. &
a plushie \{label\}. \\
a origami \{label\}. &
art of a \{label\}. &
graffiti of a \{label\}. &
a drawing of a \{label\}. \\
a doodle of a \{label\}. &
a photo of the \{label\}. &
a painting of the \{label\}.&
the plastic \{label\}. \\
a sculpture of the \{label\}.&
a sketch of the \{label\}.&
a tattoo of the \{label\}.&
the toy \{label\}. \\
a rendition of the \{label\}.&
the embroidered \{label\}.&
the cartoon \{label\}.&
the \{label\} in a video game. \\
the plushie \{label\}.&
the origami \{label\}.&
art of the \{label\}.&
graffiti of the \{label\}. \\
a drawing of the \{label\}.&
a doodle of the \{label\}.& & \\
\midrule
\multicolumn{4}{l}{\bf Stanford Cars} \\
a photo of a \{label\}.&
a photo of the \{label\}.&
a photo of my \{label\}.&
i love my \{label\}! \\
a photo of my dirty \{label\}.&
a photo of my clean \{label\}.&
a photo of my new \{label\}.&
a photo of my old \{label\}. \\
\midrule
\multicolumn{4}{l}{\bf DTD} \\
a photo of a \{label\} texture.&
a photo of a \{label\} pattern.&
a photo of a \{label\} thing.&
a photo of a \{label\} object. \\
a photo of the \{label\} texture. &
a photo of the \{label\} pattern. &
a photo of the \{label\} thing. &
a photo of the \{label\} object. \\
\midrule
\multicolumn{4}{l}{\bf FGVC Aircraft} \\
a photo of a \{label\}, a type of aircraft.&
a photo of the \{label\}, a type of aircraft.& & \\
\midrule
\multicolumn{4}{l}{\bf Flowers102} \\
a photo of a \{label\}, a type of flower. &&& \\
\midrule
\multicolumn{4}{l}{\bf Pets } \\
a photo of a \{label\}, a type of pet.&&& \\
\midrule
\multicolumn{4}{l}{\bf  SUN39} \\
a photo of a \{label\}.&
a photo of the \{label\}.&& \\
\midrule
\multicolumn{4}{l}{\bf  ImageNet} \\
a bad photo of a \{label\}. & 
a photo of many \{label\}. &
a sculpture of a \{label\}. &
a photo of the hard to see \{label\}. \\
a low resolution photo of the \{label\}. & 
a rendering of a \{label\}. &
graffiti of a \{label\}. &
a bad photo of the \{label\}.  \\
a cropped photo of the \{label\}. &
a tattoo of a \{label\}. & 
the embroidered \{label\}. &
a photo of a hard to see \{label\}.  \\
a bright photo of a \{label\}.&
a photo of a clean \{label\}.&
a photo of a dirty \{label\}.&
a dark photo of the \{label\}. \\
a drawing of a \{label\}.&
a photo of my \{label\}.&
the plastic \{label\}.&
a photo of the cool \{label\}. \\
a close-up photo of a \{label\}.&
a black and white photo of the \{label\}.&
a painting of the \{label\}.&
a painting of a \{label\}. \\
a pixelated photo of the \{label\}.& 
a sculpture of the \{label\}.&
a bright photo of the \{label\}.&
a cropped photo of a \{label\}. \\
a plastic \{label\}.&
a photo of the dirty \{label\}.& 
a jpeg corrupted photo of a \{label\}.&
a blurry photo of the \{label\}. \\
a photo of the \{label\}.&
a good photo of the \{label\}.&
a rendering of the \{label\}.&
a \{label\} in a video game. \\
a photo of one \{label\}.&
a doodle of a \{label\}.&
a close-up photo of the \{label\}.&
a photo of a \{label\}. \\
the origami \{label\}.&
the \{label\} in a video game.&
a sketch of a \{label\}.&
a doodle of the \{label\}. \\
a origami \{label\}.&
a low resolution photo of a \{label\}.&
the toy \{label\}.&
a rendition of the \{label\}. \\
a photo of the clean \{label\}.& 
a photo of a large \{label\}.& 
a rendition of a \{label\}.&
a photo of a nice \{label\}. \\
a photo of a weird \{label\}.& 
a blurry photo of a \{label\}.&
a cartoon \{label\}.&
art of a \{label\}. \\
a sketch of the \{label\}.& 
a embroidered \{label\}.&
a pixelated photo of a \{label\}.&
itap of the \{label\}. \\
a jpeg corrupted photo of the \{label\}.& 
a good photo of a \{label\}.&
a plushie \{label\}.&
a photo of the nice \{label\}. \\
a photo of the small \{label\}.& 
a photo of the weird \{label\}.&
the cartoon \{label\}.&
art of the \{label\}. \\
a drawing of the \{label\}.& 
a photo of the large \{label\}.& 
a black and white photo of a \{label\}.&
the plushie \{label\}. \\
a dark photo of a \{label\}.& 
itap of a \{label\}.& 
graffiti of the \{label\}.& 
a toy \{label\}. \\
itap of my \{label\}.& 
a photo of a cool \{label\}.&
a photo of a small \{label\}.& 
a tattoo of the \{label\}. \\
\bottomrule
\end{tabular}}
\caption{The list of prompts used to evaluate the performance of zero-shot classification on 11 visual recognition datasets.}
\label{tab:prompt}
\end{table*}